%% file: dx.tex
\documentclass[a4paper]{article}
\usepackage[utf8]{inputenc}
\usepackage[T1]{fontenc}
\usepackage{ae,aecompl}
\usepackage{amsfonts}
\usepackage{amsmath}
\usepackage{amsthm}

\usepackage{times}

\usepackage{graphicx}
\usepackage{xcolor}

\graphicspath{{./figures/}}

\definecolor{ourbrown}{RGB}{155,100,15}
\definecolor{ourpurple}{RGB}{145,0,140}
\definecolor{darkgreen}{RGB}{0,170,0}
\definecolor{darkorange}{RGB}{225,100,0}

\newtheorem{assumption}{Assumption}

\title{LSTM for Model-Based Anomaly Detection in Cyber-Physical Systems}
\author%
{%
Benedikt Eiteneuer$^1$ \and 
Oliver Niggemann$^1$\\
$^1$Institute Industrial IT, OWL University of Applied Sciences \\
e-mail: \{benedikt.eiteneuer,oliver.niggemann\}@hs-owl.de\\
}

\date{August 2018\\
Proceedings of the 29th International Workshop on Principles of Diagnosis DX-2018, Warsaw, Poland}

\begin{document}

\maketitle

\input{abstract}
\input{introduction}

\input{method2}
\input{experiments}
\input{conclusion}

\section*{Acknowledgments}
The authors thank Nemanja Hranisavljevic for helpful discussions and remarks.
This work has received funding from the European Unions Horizon 2020 research and innovation programme under grant agreement No. 678867.

\bibliographystyle{unsrt}
\bibliography{references}

\end{document}

%% file: abstract.tex

\begin{abstract}
Anomaly detection is the task of detecting data which differs from the normal behaviour of a system in a given context.
In order to approach this problem, data-driven models can be learned to predict current or future observations.
Oftentimes, anomalous behaviour depends on the internal dynamics of the system and looks normal in a static context. To address this problem, the model should also operate depending on state.
Long Short-Term Memory (LSTM) neural networks have been shown to be particularly useful to learn time sequences with varying length of temporal dependencies
and are therefore an interesting general purpose approach to learn the behaviour of arbitrarily complex Cyber-Physical Systems.

In order to perform anomaly detection, we slightly modify the standard norm 2 error to incorporate an estimate of model uncertainty. 
We analyse the approach on artificial and real data.

\end{abstract}

%% file: introduction.tex
\section{Introduction}

The diagnosis of time-dependent systems has always been a focus of research in domains such as industrial applications, robotics, medical diagnosis and many more \cite{gao2015survey, niggemann2012learning}.
Typical applications of diagnosis involve the detection of anomalous system behaviour, system degradation or sub-optimal conditions concerning system health, energy consumption or product quality.

Traditionally, manual models or simulations based on the system description and its physics have been created by experts of their domain.
However, the ever-increasing complexity of distributed multi-component Cyber-Physical Systems (CPS) have made these manual efforts increasingly difficult. Automated data-driven machine learning applications fill in the need of model creation to describe the system behaviour {\cite{niggemann2015diagnosis}.
Because in many such systems a safe and reliable functioning of its components is a vital requirement, it is important to detect anomalous or faulty system behaviour in real time and as early as possible.

In general, a data sample can be considered anomalous if its generating distribution is significantly different from the normal behaviour. Usually, such anomalies are not observed abundantly on real systems because it is too expensive and sometimes even dangerous to record such data. Furthermore, even if one is able to record (and label!) anomalous data properly, it would still not be sufficiently exhaustive for a classification task, because anomalies are fundamentally diverse.
Instead, one typically assumes a (sub)set of recorded data to not contain any anomalies, thereby labelling all training data as \textit{normal}, called one-class-classification \cite{khan2014one}.
The training data is then used to learn a model of the normal behaviour.
When doing inference, new data is compared to the expectation and then classified, see \cite{pimentel2014review} for a recent review of common such methods.

Many commonly used techniques do not naturally make use of the time ordering
and are therefore unable to detect certain kind of temporal anomalies which depend on the systems internal state.
Methods that try to adapt commonly used approaches into the time domain, such as sliding windows usually only model fixed time temporal dependencies, while methods analysing spectral properties such as FFT rely on proper periodic lengths. 

Long Short-Term Memory \cite{doi:10.1162/neco.1997.9.8.1735} (LSTM) networks are a natural candidate to fill in this gap because they keep a memory of its past input history and are theoretically able to learn arbitrary lengths of temporal patterns. These internal, long-term memory states can be shaped by the learning algorithm to be useful in order to make present and future predictions for anomaly detection.
LSTM have been used for exactly this purpose, see e.g. \cite{wielgosz2017using, o2016recurrent, nanduri2016anomaly}.

However, for the purpose of anomaly detection, it is not sufficient to compare a model's predictions to the actual values. In order to assess how much to trust the model, an estimate of uncertainty is necessary.

The authors of \cite{malhotra2015long} approach this problem by fitting a multivariate normal distribution, \cite{fortunato2017bayesian, solch2016variational} use variational inference while \cite{zhu2017deep} use a Monte Carlo dropout scheme to handle model uncertainty.

In this paper we go a more direct route by learning explicit uncertainty predictions. We make the following contributions:
\begin{itemize}
    \item Modify and motivate the standard loss function to incorporate estimates of noise inherent to the data. The problem of anomaly detection can only be tackled if anomaly scores can be directly related to probabilities of events given some assumptions.
    \item Demonstrate the advantages of modelling internal state dynamics in order to properly monitor dynamic systems.
    \item Analyse the learned state representation of LSTM in CPS data with respect to anomalous data.
\end{itemize}

The Paper is structured as follows:
In Section \ref{sec:method}, we start off by describing the general features of CPS that are important to the problem. We then derive a modified loss function which enables a model to \textit{predict the noise along with the data}. After a brief description of the LSTM model, we describe the anomaly detection approach by incorporating the prediction uncertainty into the anomaly classification task.
In Section \ref{sec:experiments}, we analyse the approach with a simple artificial level control system and a real-world power consumption time series data set.
Discussion and some general remarks are given in Section \ref{sec:discussion}. We conclude in Section \ref{sec:conclusion}.

%% file: method2.tex
\section{Method}\label{sec:method}


\subsection{Problem Description}\label{subsec:problem}
We define a CPS according to the following statements:
\begin{itemize}
\item A CPS is an isolated, dynamic system which interacts with its environment via signal inputs $\mathbf{x}_t\in\mathcal{X}\subseteq\mathbb{R}^p$ and outputs $\mathbf{y}_t\in\mathcal{Y}\subseteq\mathbb{R}^q$, where $t$ denotes time, $p$ and $q$ in- and output dimensions repectively. Amongst others, a CPS can be comprised of mechanical, electrical or biological parts.
\item Although the underlying system is continuous, all system in- and output is given in discrete time steps.
\item At any given point in time, the CPS has an internal state $\mathbf{s}_t\in\mathcal{S}\subseteq\mathbb{R}^n$. A state is the minimal description needed to theoretically be able to predict the outputs $\mathbf{y}_{t}$ as well as the next state, if only the current input $\mathbf{x}_t$ is given in addition.
\item The system inputs $\mathbf{x}$ can indicate discrete or continuous events triggering a state change. These can be external or internal influences which have an environmental, cybernetic (e.g. actor signals), human or even random origin.
\item The system outputs $\mathbf{y}$ are observations taken on the system that depend both on its internal state and its input. In general, such observations could be given by some arbitrary non-linear function $f(\mathbf{s},\mathbf{x})$.
\item The observation vector $\mathbf{y}$ is subject to noise and can therefore only be determined up to some leftover covariance, $V_t\in\mathbb{R}^{q\times q}$
\end{itemize}

\begin{figure}
    \centering
    \def\svgwidth{0.95\columnwidth}
    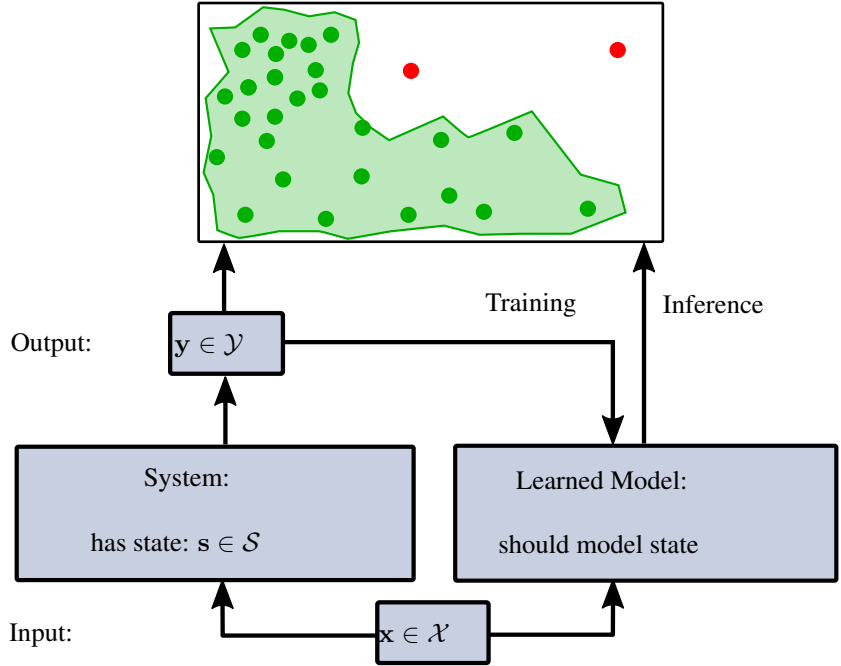
    \caption{Sketch of the work flow. The system in- and output is used to learn a model of expected outputs $\mathbf{y}$. When doing inference, the prediction is compared to the actual output in order to detect anomalies.}
    \label{fig:system}
\end{figure}

In order to monitor the system behaviour, the vector of measured observables $\mathbf{y}_t$ has to be learned for each step in time $t$.
It can then be compared to the actual value and information about the system health, such as degradation and anomalous or untypical behaviour, can be inferred.
Because $V_t$ is not known a priori and its knowledge is important to the anomaly classification task, \textit{its determination is part of the learning process}.

If complete information of a deterministic CPS is available in the $\mathbf{x}$ and $\mathbf{s}$ variables, it is possible to find a mapping $f:(\mathbf{s}_{t-1},\mathbf{x}_t)\mapsto (\mathbf{y}_t, V_t$).
However, the internal state of the system is not actually measured, but only the vector of observations $\mathbf{y}$. Therefore, the $\mathbf{s}_t$ variable has to be modelled along with the data by inference of the system behaviour in time as well as the inter-dependencies of the $\mathbf{x}_t$ and $\mathbf{y}_t$ variables, see Figure \ref{fig:system}.

\subsection{Loss Function}\label{subsec:lstm}
A simple feed-forward neural net can approximate any function \cite{hornik1989multilayer} and is therefore a good candidate to find the desired mapping $f$, if \textit{all information} were given along with $\mathbf{x}$ and the system had no internal state.
If this is not the case, a state can be modelled within a neural network by feeding information about past input through time, which is exactly what a recurrent architecture does.
LSTM's are particularly equipped for this task, because a gating mechanism allows gradients to flow back in time up to arbitrary lengths, enabling the learning of diverse temporal dependencies. We would like to remind the reader, that if those temporal patterns are well known (and maybe not too long) it is not necessary to learn such states. In cases such as these we could just group the input up to the desired point back in time and analyse the aggregate, e.g.\ with a simple neural network. The strength of learning internal state representations lies in no prior knowledge given about possible temporal dependencies.

The problem is now to find a mapping of the form:
$$
    f: (\mathbf{s}_{t-1}, \mathbf{x}_t) \mapsto (\mathbf{s}_{t}, \mathbf{y}_t, V_t).
$$

\begin{assumption}
It is possible to model these states. This means that all relevant information for the prediction of outputs $\textbf{y}_t$ can be inferred from the dynamic behaviour.
\end{assumption}
It is then possible to predict $\mathbf{y}_t$ up to some leftover statistical uncertainty. If this were not the case, missing information could only be compensated for with additional in- or output variables or by adding more data.
If the assumption can not be met, there will be a number of unknown internal states that are not modelled properly. System behaviour that depends on these \textit{hidden} states can then not properly be accounted for.

We denote the model prediction with $\hat{\mathbf{y}}_t(\mathbf{s}_{t-1},\mathbf{x}_t|\theta)$, $\theta$ being the model parameter.
\begin{assumption}
The leftover covariance of model-to-actual difference $\hat{\mathbf{y}}_t-\mathbf{y}_t$ is Gaussian. 
\label{ass:variance}\end{assumption}
This is motivated by the idea of including as much information as possible in the system state and the ability to propagate this information into the model outputs, which leaves maximum entropy in the prediction difference. Note that the normal distribution maximises the entropy for a real valued random variable with specified mean and variance.
The whole process can be understood similar to a multivariate Gaussian random variable in a Gaussian process with mean $\mathbf{\hat{y}}_t$ and white noise, meaning:
$\mathrm{cov}(\mathbf{y}_t,\mathbf{y}_{t'})=\delta_{tt'}V_t$

$\mathbf{y}_t$ is then distributed normally with mean $\hat{\mathbf{y}}_t$ and some covariance matrix $V_t(\mathbf{s}_{t-1},\mathbf{x}_t|\theta)$:
\begin{equation}
\begin{aligned}
    p_\theta(\mathbf{y}_t|\mathbf{s}_{t-1}, & \mathbf{x}_t)= \left(2\pi\right)^{-\frac{q}{2}}\left|V_t\right|^{-\frac{1}{2}}\times\\
    & \exp\left[-\frac{1}{2}\left(\mathbf{y}_t-\hat{\mathbf{y}}_t\right)^T V^{-1}_t\left(\mathbf{y}_t-\hat{\mathbf{y}}_t\right)\right].
\end{aligned}
\end{equation}

\begin{assumption}
All elements of $y_t^{i}\in \mathbf{y}_t$ are conditionally independent from each other given $\mathbf{x}_t$ and $\mathbf{s}_{t-1}$.
\end{assumption}
In a physical system, this corresponds to the fact, that two sensors might take measurements within the same environment and therefore are (possibly maximally) correlated, but given the \textit{complete} internal state of the system, the remaining noise on each sensor is independent from each other. (They are two \textit{different} sensors.)

$V_t$ is therefore diagonal and $p_\theta(\mathbf{y}_t|\mathbf{s}_{t-1},\mathbf{x}_t)$ factorises to the following expression:
\begin{equation}
    p_\theta(\mathbf{y}_t|\mathbf{s}_{t-1},\mathbf{x}_t)=\prod_{i=0}^{q}\frac{1}{\sqrt{2\pi}\sigma_t^i}\exp\left[-\frac{1}{2}\left(\frac{\mathbf{y}_t^i-\mathbf{\hat{y}}_t^i}{\mathbf{\sigma}_t^i}\right)^2\right],\label{eq:likelihood}
\end{equation}
where $\sigma_t^i$ are the diagonal elements of $V_t$ and can be summarised in the vector $\mathbf{\sigma}_t$.

Maximising the likelihood function (\ref{eq:likelihood}) is equivalent to minimising the negative log likelihood loss function:
\begin{equation}
    L_t=\sum_i\left[\left(\frac{\mathbf{y}_t^i-\mathbf{\hat{y}}_t^i}{\sigma_t^i}\right)^2+2\log\sigma_t^i\right]
\end{equation}



\subsection{Long-Short Term Memory Network}
We will employ a vanilla LSTM \cite{doi:10.1162/neco.1997.9.8.1735} network to find $f$ and model all of the $\mathbf{\hat{y}}_t$ and  $\sigma_t$ (as its short-term memory state) \emph{as well} as the hidden state $\mathbf{s}_t$ (as its long-term memory states) and use supervised learning with input $\mathbf{x}_t$ on the targets $\mathbf{y}_t$ with stochastic gradient descent to train the weights of the parametric function.

\begin{figure}[tb]
    \centering
    \def\svgwidth{1.0\columnwidth}
    \begin{footnotesize}
    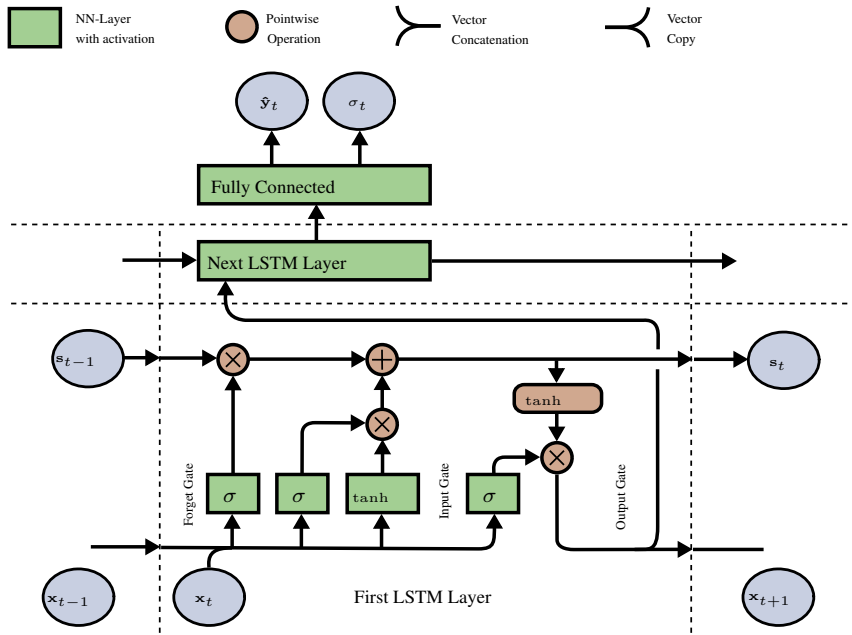
    \end{footnotesize}
    \caption{LSTM architecture.}
    \label{fig:lstm_architecture}
\end{figure}

Writing $\sigma_t=\exp(\tau_t)$ guarantees $\sigma>0$ and provides numerical stability in the learning process.
The network architecture consists of several (1-5) fully recurrent, stacked LSTM layers with a subsequent fully connected tanh activation neural network with $2q$ output channels, one for each $y^i$ and $\tau^i$, see Figure \ref{fig:lstm_architecture}. Gradients are truncated in time after $t_{\mathrm{max}}$ steps, typically between 5 and 50.
Optimised is the sum over $L_t$ containing all terms from the current time $t$ to $t-t_{\mathrm{max}}$ steps into the past. The training can be realised in a mini-batch process by going through the data sequences in parallel.

The models used for this work are implemented within the \textsc{TensorFlow} \cite{abadi2016tensorflow} machine learning framework.

\begin{figure}[tb]
    \centering
    \includegraphics[width=1.0\columnwidth]{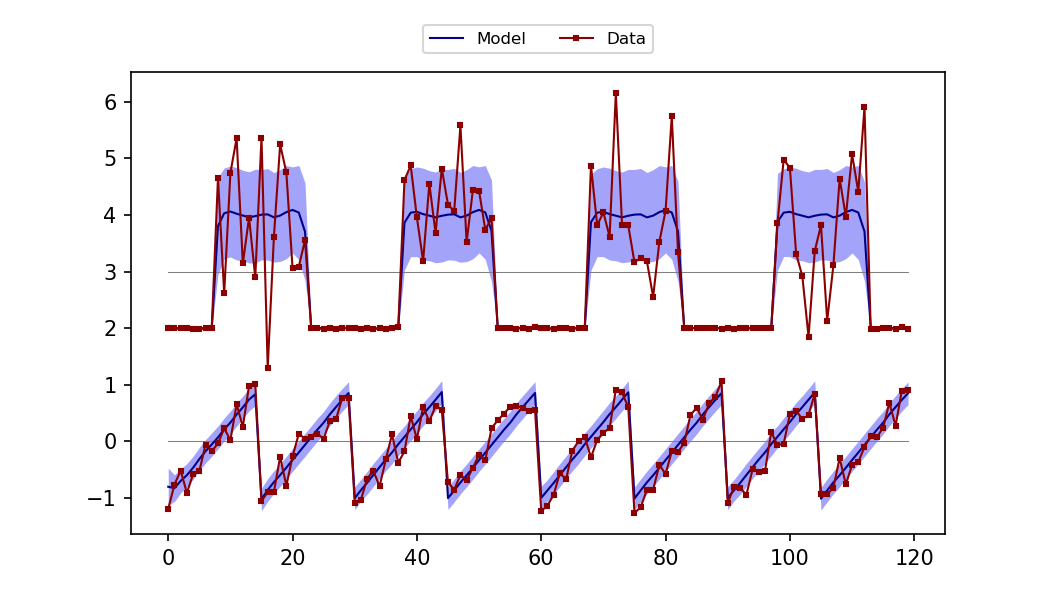}
    \caption{Two periodic, artificial signals learned jointly with an LSTM network. The signal $\mathbf{y}$ (red) is sampled with different noise levels in different domains, first gaussian and second uniform. The LSTM is able to learn both mean $\mathbf{\hat{y}}$ (blue) as well as standard deviation $\sigma$ (shaded 1-$\sigma$). The input $\mathbf{x}$ is always zero except once at the start of each period (0, 30, ..) where input equals one.}
    \label{fig:sim_signals}
\end{figure}

In the learning stage the neural network will optimise its parameters in order to model the domain dependent functions $\mathbf{\hat{y}}$ and $\sigma$ as accurate as possible, see Figure \ref{fig:sim_signals}. Put differently, the neural net learns the \textit{domain mean}, taking for each known state the average over the known observations falling into that category. The method is therefore self regularising as long as enough data exists for \textit{each domain} to determine sensible mean and standard deviation values.
While this can be said for any sufficiently simple model, the strength of this approach lies in the learning and grouping of temporal patterns into states which are forced to be treated \textit{unified} within the model.

\subsection{Anomaly Detection and System degradation}\label{subsec:anomaly}

One advantage of modelling the noise along with the system output can be understood in the first signal of the Figure \ref{fig:sim_signals}. Without learning noisy patterns, one would be far less sensitive to deviations of order 1 in the regimes with little to zero noise, because the same signal exhibits a noise level of the same order of magnitude in another regime. By modelling the noise, anomaly detection can be performed for every observed signal independently of different system behaviour (through time).

\begin{figure}[tb]
    \centering
    \includegraphics[width=1.00\columnwidth]{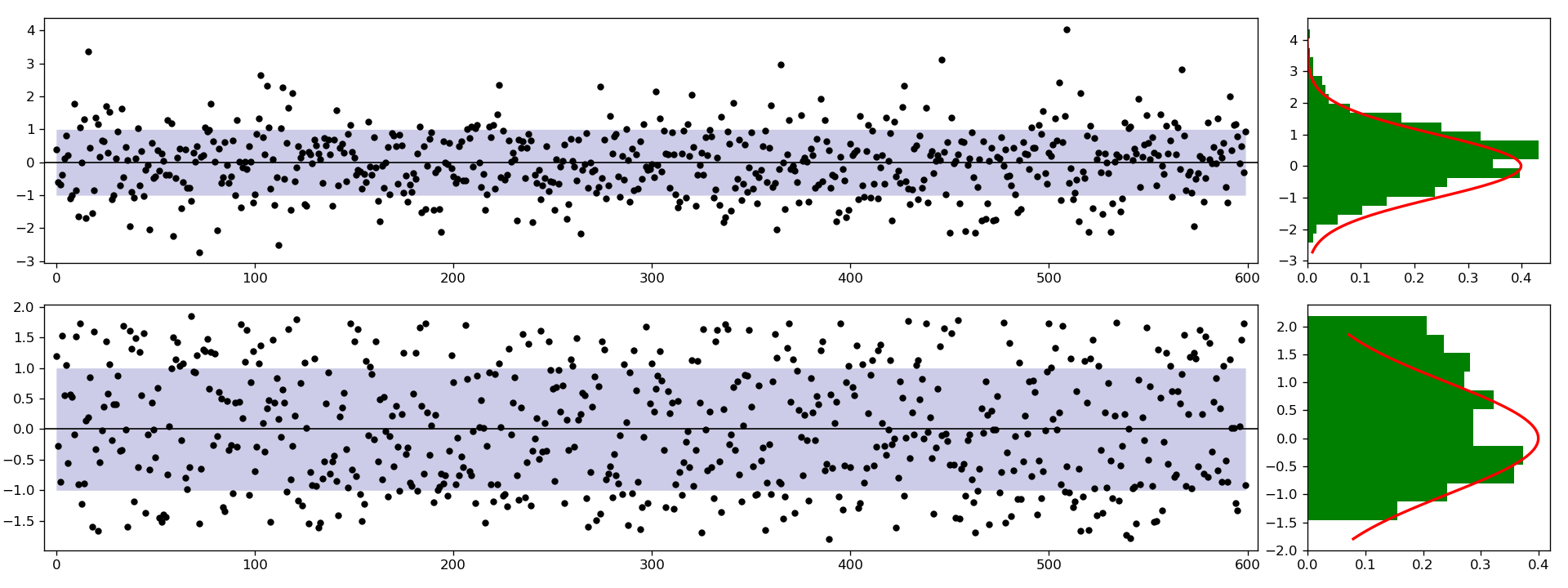}
    \caption{Residuals $(\mathbf{\hat{y}}-\mathbf{y})/\sigma$ for the two examplary periodic functions shown in Figure \ref{fig:sim_signals}. The first signal's noise is sampled from a gaussian with different variance at different domains in the cycle while the second signal is sampled uniformly. On the right, the residual distribution is histogrammed and compared to a normal distribution.}
    \label{fig:sim_signals_res}
\end{figure}
To discriminate between normal and anomalous data, we can calculate the normalised residuals $r_t=(\mathbf{\hat{y}}_t-\mathbf{y}_t)/\sigma_t$ which should be distributed normally:
\begin{equation}
    p(r_t|s_{t-1},x_t)=\mathcal{N}(r_t;0,1),
\end{equation}

Figure \ref{fig:sim_signals_res} shows the graph of the residuals from Figure \ref{fig:sim_signals} as well as a histogram of its distribution. The histogram of the first signal (noise with 2 modal variance) resembles a normal distribution and shows that the LSTM correctly approximates the different noise levels depending on the current progress in the period. The second signal has a uniform noise, which cannot be modelled perfectly. Strictly speaking, the Assumption \ref{ass:variance} is not fullfilled in this case. However, the variance can still be learned. As is clear for the purpose of anomaly detection, the only important issue is that residual outliers should be similarly sparse in the \textit{tails} of the distribution. Most problematic cases would be data-distributions with very long/flat tails which could result in false positive classifications because the learned variance is a bad measure for the decision boundary.

When should samples be considered as anomalous?
Generally, it is possible to combine several samples and/or features and then analyse the aggregate statistics. A simple method is to threshold the normal distribution given the test point $r_t$. Such a threshold can either be learned (as is often done in the literature \cite{pimentel2014review}) on the training or held out data to include a specified percentile of the training data (e.g. 95-100\%) or calculated to include a specified percentile of the expected distribution.

If a percentile is calculated from the normal distribution (e.g. $r=3$ corresponds to a $1-2.7\times10^{-3}$ quantile), this is the expected frequency of points to be classified as anomalous (False positive rate). Of course, anomalous points themselves are not expected to be distributed normally and thus should exceed the threshold value much more frequently.

The strength of this approach is that it is not necessary to learn the threshold with held out data, possibly containing known anomalies. Instead, it is not necessary at all to have ever measured an anomaly, which is how it should be. Approximate statements about the expected frequency of false positives can be made and thresholds can be chosen according to ones preferences given the size of the test data and desired sensitivity.


In order to monitor system change over time, again the residual statistics can be used. In a typical degradation scenario at least one of the observables usually exhibits more and more anomalous behaviour.

%% file: figures/system+AD.pdf_tex
\begingroup%
  \makeatletter%
  \providecommand\color[2][]{%
    \errmessage{(Inkscape) Color is used for the text in Inkscape, but the package 'color.sty' is not loaded}%
    \renewcommand\color[2][]{}%
  }%
  \providecommand\transparent[1]{%
    \errmessage{(Inkscape) Transparency is used (non-zero) for the text in Inkscape, but the package 'transparent.sty' is not loaded}%
    \renewcommand\transparent[1]{}%
  }%
  \providecommand\rotatebox[2]{#2}%
  \ifx\svgwidth\undefined%
    \setlength{\unitlength}{230.08129445bp}%
    \ifx\svgscale\undefined%
      \relax%
    \else%
      \setlength{\unitlength}{\unitlength * \real{\svgscale}}%
    \fi%
  \else%
    \setlength{\unitlength}{\svgwidth}%
  \fi%
  \global\let\svgwidth\undefined%
  \global\let\svgscale\undefined%
  \makeatother%
  \begin{picture}(1,0.76366713)%
    \put(0,0){\includegraphics[width=\unitlength,page=1]{system+AD.pdf}}%
    \put(-0.00290743,0.02808946){\color[rgb]{0,0,0}\makebox(0,0)[lb]{\smash{Input:}}}%
    \put(0.41867526,0.02673973){\color[rgb]{0,0,0}\makebox(0,0)[lb]{\smash{$\mathbf{x}\in\mathcal{X}$}}}%
    \put(0.15093386,0.20638351){\color[rgb]{0,0,0}\makebox(0,0)[lb]{\smash{System:}}}%
    \put(0.5560355,0.12956523){\color[rgb]{0,0,0}\makebox(0,0)[lb]{\smash{should model state}}}%
    \put(0,0){\includegraphics[width=\unitlength,page=2]{system+AD.pdf}}%
    \put(-0.00106207,0.36164428){\color[rgb]{0,0,0}\makebox(0,0)[lb]{\smash{Output:}}}%
    \put(0.1858978,0.36037733){\color[rgb]{0,0,0}\makebox(0,0)[lb]{\smash{$\mathbf{y}\in\mathcal{Y}$}}}%
    \put(0.57638517,0.20257768){\color[rgb]{0,0,0}\makebox(0,0)[lb]{\smash{Learned Model:}}}%
    \put(0,0){\includegraphics[width=\unitlength,page=3]{system+AD.pdf}}%
    \put(0.54123528,0.40761113){\color[rgb]{0,0,0}\makebox(0,0)[lb]{\smash{Training}}}%
    \put(0.74392063,0.40526249){\color[rgb]{0,0,0}\makebox(0,0)[lb]{\smash{Inference}}}%
    \put(0,0){\includegraphics[width=\unitlength,page=4]{system+AD.pdf}}%
    \put(0.09047206,0.13085917){\color[rgb]{0,0,0}\makebox(0,0)[lb]{\smash{has state: $\mathbf{s}\in\mathcal{S}$}}}%
  \end{picture}%
\endgroup%

%% file: lstm.pdf_tex
\begingroup%
  \makeatletter%
  \providecommand\color[2][]{%
    \errmessage{(Inkscape) Color is used for the text in Inkscape, but the package 'color.sty' is not loaded}%
    \renewcommand\color[2][]{}%
  }%
  \providecommand\transparent[1]{%
    \errmessage{(Inkscape) Transparency is used (non-zero) for the text in Inkscape, but the package 'transparent.sty' is not loaded}%
    \renewcommand\transparent[1]{}%
  }%
  \providecommand\rotatebox[2]{#2}%
  \ifx\svgwidth\undefined%
    \setlength{\unitlength}{372.53608722bp}%
    \ifx\svgscale\undefined%
      \relax%
    \else%
      \setlength{\unitlength}{\unitlength * \real{\svgscale}}%
    \fi%
  \else%
    \setlength{\unitlength}{\svgwidth}%
  \fi%
  \global\let\svgwidth\undefined%
  \global\let\svgscale\undefined%
  \makeatother%
  \begin{picture}(1,0.685763)%
    \put(0,0){\includegraphics[width=\unitlength,page=1]{lstm.pdf}}%
    \put(0.07410995,0.66657663){\color[rgb]{0,0,0}\makebox(0,0)[lb]{\smash{\tiny{NN-Layer}}}}%
    \put(0.2807421,0.66657663){\color[rgb]{0,0,0}\makebox(0,0)[lb]{\smash{\tiny{Pointwise}}}}%
    \put(0.48208708,0.66585762){\color[rgb]{0,0,0}\makebox(0,0)[lb]{\smash{\tiny{Vector}}}}%
    \put(0.233799,0.14282022){\color[rgb]{0,0,0}\makebox(0,0)[lb]{\smash{\footnotesize{$\sigma$}}}}%
    \put(0.47861463,0.12421891){\color[rgb]{0,0,0}\rotatebox{90}{\makebox(0,0)[lb]{\smash{\tiny{Input Gate}}}}}%
    \put(0.2010201,0.11679719){\color[rgb]{0,0,0}\rotatebox{90}{\makebox(0,0)[lb]{\smash{\tiny{Forget Gate}}}}}%
    \put(0.67009565,0.11290434){\color[rgb]{0,0,0}\rotatebox{90}{\makebox(0,0)[lb]{\smash{\tiny{Output Gate}}}}}%
    \put(0.07410995,0.6445103){\color[rgb]{0,0,0}\makebox(0,0)[lb]{\smash{\tiny{with activation}}}}%
    \put(0.48197498,0.64379128){\color[rgb]{0,0,0}\makebox(0,0)[lb]{\smash{\tiny{Concatenation}}}}%
    \put(0.30894043,0.14281784){\color[rgb]{0,0,0}\makebox(0,0)[lb]{\smash{\footnotesize{$\sigma$}}}}%
    \put(0.37417399,0.14274574){\color[rgb]{0,0,0}\makebox(0,0)[lb]{\smash{\tiny{$\tanh$}}}}%
    \put(0.51583052,0.14281784){\color[rgb]{0,0,0}\makebox(0,0)[lb]{\smash{\footnotesize{$\sigma$}}}}%
    \put(0.20304961,0.03341943){\color[rgb]{0,0,0}\makebox(0,0)[lb]{\smash{\tiny{$\mathbf{x}_t$}}}}%
    \put(0.80396662,0.03547489){\color[rgb]{0,0,0}\makebox(0,0)[lb]{\smash{\tiny{$\mathbf{x}_{t+1}$}}}}%
    \put(0.04300251,0.03403682){\color[rgb]{0,0,0}\makebox(0,0)[lb]{\smash{\tiny{$\mathbf{x}_{t-1}$}}}}%
    \put(0.05234964,0.29611797){\color[rgb]{0,0,0}\makebox(0,0)[lb]{\smash{\tiny{$\mathbf{s}_{t-1}$}}}}%
    \put(0.8266278,0.29186946){\color[rgb]{0,0,0}\makebox(0,0)[lb]{\smash{\tiny{$\mathbf{s}_{t}$}}}}%
    \put(0.28186176,0.6445103){\color[rgb]{0,0,0}\makebox(0,0)[lb]{\smash{\tiny{Operation}}}}%
    \put(0.71568083,0.66657663){\color[rgb]{0,0,0}\makebox(0,0)[lb]{\smash{\tiny{Vector}}}}%
    \put(0.71556879,0.6445103){\color[rgb]{0,0,0}\makebox(0,0)[lb]{\smash{\tiny{Copy}}}}%
    \put(0.56152487,0.24944038){\color[rgb]{0,0,0}\makebox(0,0)[lb]{\smash{\tiny{$\tanh$}}}}%
    \put(0,0){\includegraphics[width=\unitlength,page=2]{lstm.pdf}}%
    \put(0.37622841,0.03216184){\color[rgb]{0,0,0}\makebox(0,0)[lb]{\smash{\scriptsize{First LSTM Layer}}}}%
    \put(0,0){\includegraphics[width=\unitlength,page=3]{lstm.pdf}}%
    \put(0.21682396,0.39821807){\color[rgb]{0,0,0}\makebox(0,0)[lb]{\smash{\scriptsize{Next LSTM Layer}}}}%
    \put(0,0){\includegraphics[width=\unitlength,page=4]{lstm.pdf}}%
    \put(0.22072741,0.48134694){\color[rgb]{0,0,0}\makebox(0,0)[lb]{\smash{\scriptsize{Fully Connected}}}}%
    \put(0,0){\includegraphics[width=\unitlength,page=5]{lstm.pdf}}%
    \put(0.27396543,0.57574219){\color[rgb]{0,0,0}\makebox(0,0)[lb]{\smash{\tiny{$\mathbf{\hat{y}}_t$}}}}%
    \put(0,0){\includegraphics[width=\unitlength,page=6]{lstm.pdf}}%
    \put(0.36924217,0.5757422){\color[rgb]{0,0,0}\makebox(0,0)[lb]{\smash{\tiny{$\sigma_t$}}}}%
    \put(0,0){\includegraphics[width=\unitlength,page=7]{lstm.pdf}}%
  \end{picture}%
\endgroup%

%% file: experiments.tex
\section{Experiments}\label{sec:experiments}

In this section, we test the approach on one artificial level control system and the power demand data set.

\input{watertank}

\input{powerdemand}


%% file: watertank.tex
\subsection{Water Tank System}\label{subsec:watertank}
\begin{figure}[htb]
    \centering
   \includegraphics[width=0.48\columnwidth]{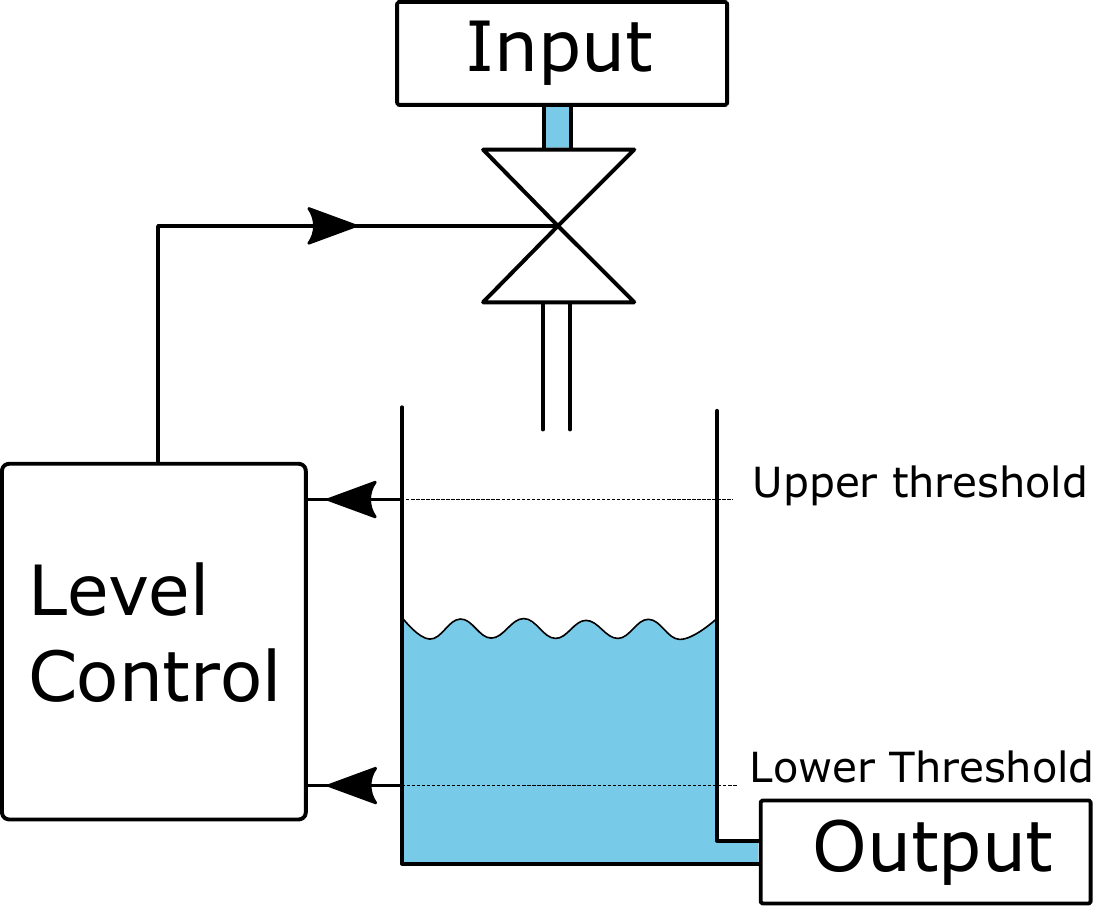}
   \setlength{\unitlength}{\columnwidth}
   \begin{picture}(0.5,0.2)
   \put(0,0){\includegraphics[width=0.48\columnwidth]{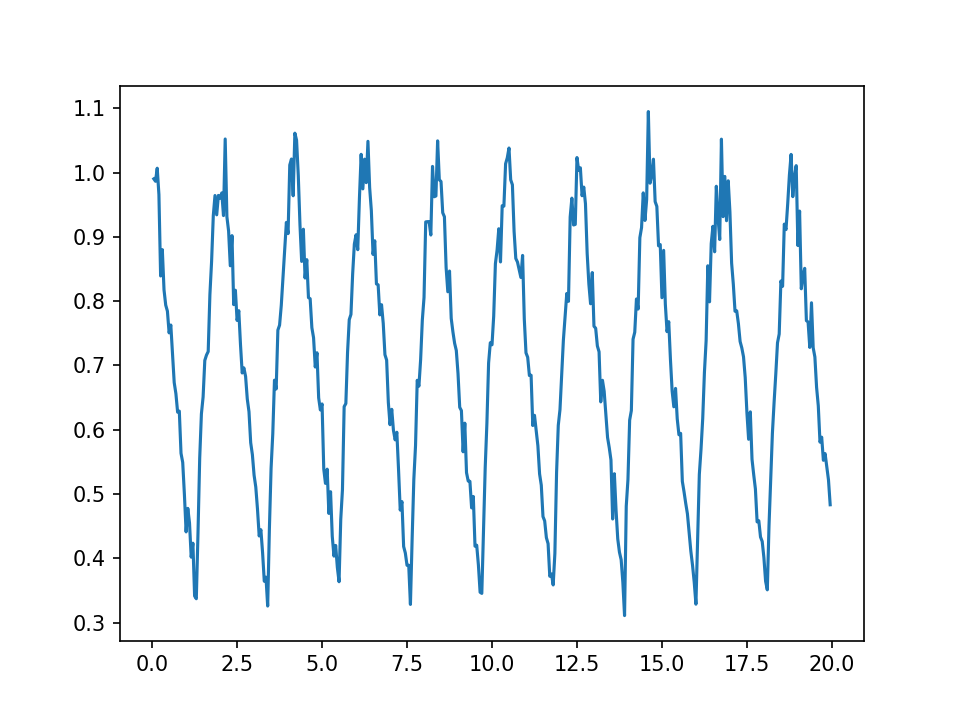}}
   \put(0.25,-0.01){\tiny{time}}
   \put(0.05,0.35){$q_{\mathrm{out}}$}
   \end{picture}
    \caption{(\textit{left}) Sketch of the water tank control system. (\textit{right}) The typical cyclic pattern of $q_\mathrm{out}$ in time.  Some noise is sampled from a Gaussian distribution with variance proportional to the absolute value.
    }
    \label{fig:watertank}
\end{figure}

The water tank system is a simple and common control system, see Figure \ref{fig:watertank}. Water is constantly flowing out of a reservoir according to the root square law which relates the output flow $q_\mathrm{out}$ to the water level $h$: $q_\mathrm{out}=a\sqrt{h}$. As soon as a certain lower threshold is reached, a valve is opened which allows the tank to be refilled, all the while it is constantly being drained from the bottom. Once the level reaches an upper threshold, the valve is closed again.

This system has an internal state (the water level) which we assume to be not measured directly as well as a clear dependency of current measurements from the past history. A certain output flux $q_{\mathrm{out},t}$ at any time will result in a larger or smaller flux in the future depending if the valve was opened or closed. The LSTM input is a binary signal about the valve's state (open/closed), while the output flux is to be predicted.

\begin{figure}
    \centering
    \includegraphics[width=0.95\columnwidth]{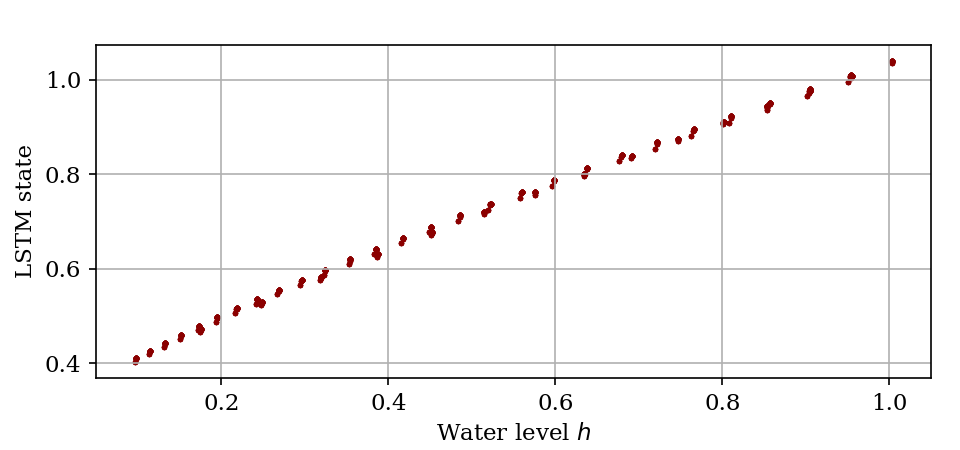}
    \caption{Scatterplot of the learned Long-Term memory state of the water tank system, given the actual water level of the system.}
    \label{fig:watertank_level_state}
\end{figure}

Figure \ref{fig:watertank_level_state} shows the learned representation of the long term memory state of the simplest LSTM architecture (with only one layer and one internal state) together with the actual water level. 
In this case, a monotonic behaviour suggest that a meaningful representation of the systems ``true" internal state has been achieved. The water level itself, which can be considered as an ideal state representation, was not part of the learning process, but only the process outflow $q_\mathrm{out}.$

\begin{figure}[htb]
    \centering
    \includegraphics[width=0.95\columnwidth]{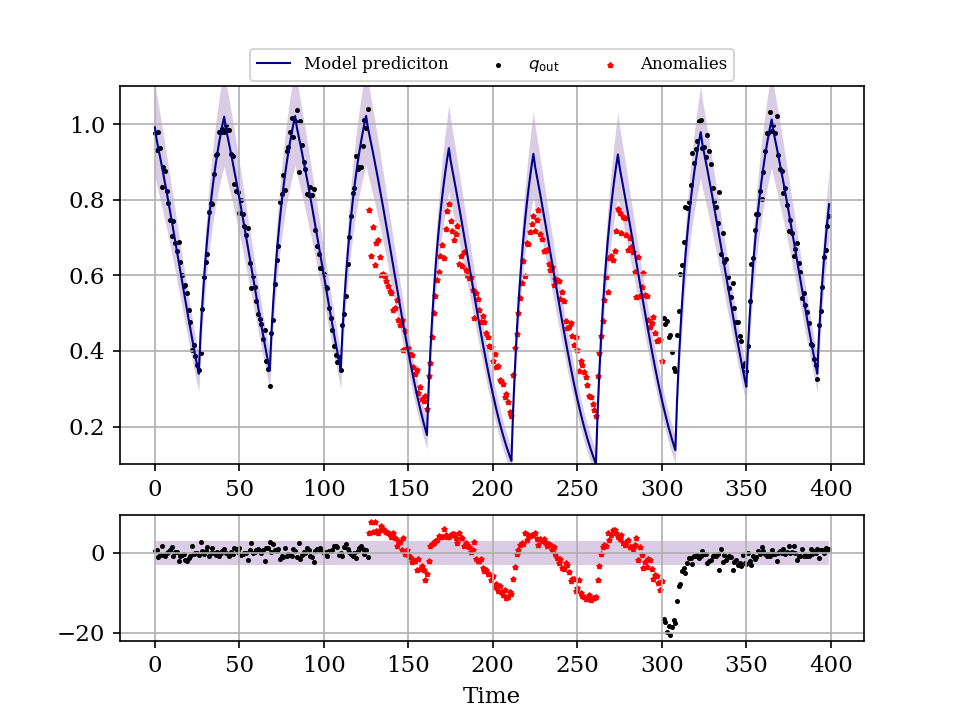}
    \caption{(\textit{Top}) Output flux of the water tank system compared to the model prediction. Shaded regions indicate $\pm3\sigma$, red points are anomalous measurements with a 25\% reduced flux. (\textit{Bottom}) Normalised residuals of the upper graph.}
    \label{fig:watertank_anom_joined}
\end{figure}

\begin{figure}[htb]
    \centering
    \includegraphics[width=0.95\columnwidth]{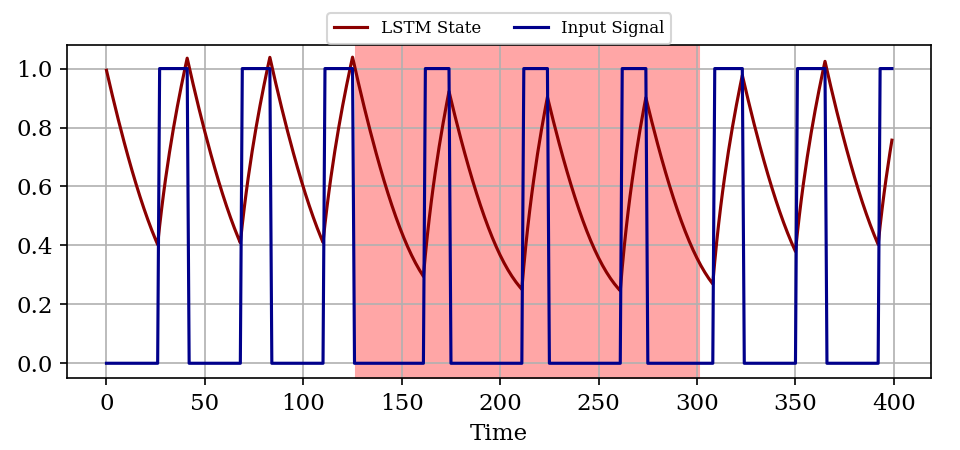}
    \caption{The internal Long-Term memory state of the water tank system that is learned by an LSTM of depth one, with only one internal state. Also shown is the input to the LSTM indicating when the valve is opened. The 75\% flux blockage is active during the red shaded period.}
    \label{fig:watertank_anom_time_state}
\end{figure}

We test the learned model against a time sequence containing a period of anomalous behaviour. To generate the anomalies, at one point a system blockage is introduced which reduces the output flux by 25\%. This kind of anomalous behaviour is not detectable with state-independent methods which do not model time in any way, as long as the flux is still large enough to be in the typical regime. Figure \ref{fig:watertank_anom_joined} shows the model prediction compared to the actual output flux.
The blockage can be detected as soon as it is introduced because observations do not match expectations immediately. Note that the first anomalous points actually conform to the usual range of observations during normal operation, and thus could not be detected by a time independent anomaly detection approach.
Figure \ref{fig:watertank_anom_time_state} shows the Long-Term memory state together with the valve control signal before, during and after the blockage. It is interesting to see, that even after the expected time for the valve to open again passed, the internal state is further reduced, matching the actual water level.

%% file: powerdemand.tex
\subsection{Power Demand}\label{subsec:power}
\begin{figure}[tb]
    \centering
    \includegraphics[width=0.95\columnwidth]{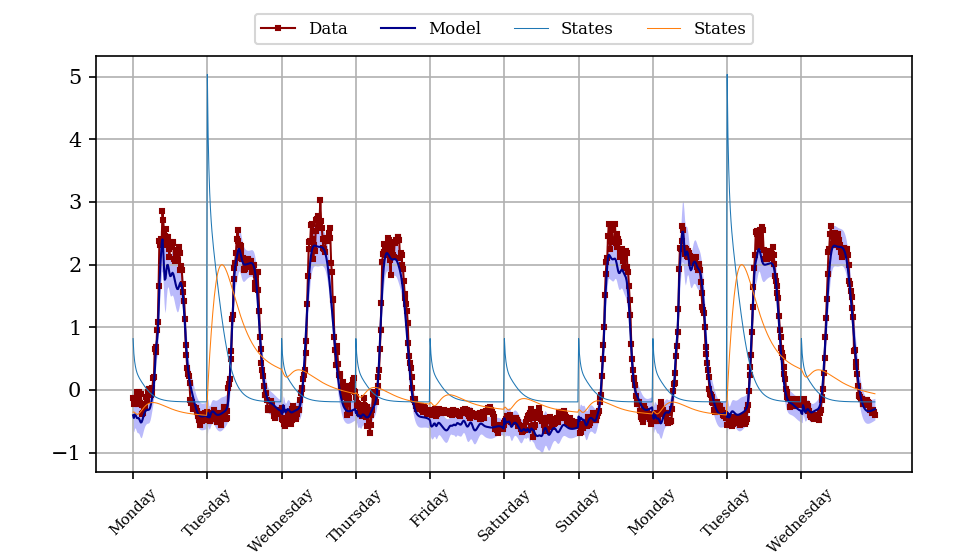}
    \includegraphics[width=0.95\columnwidth]{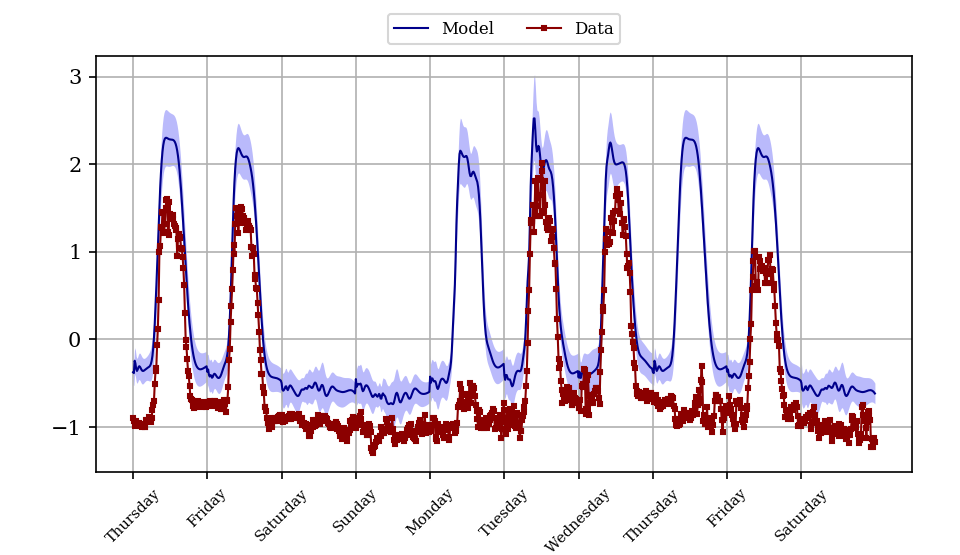}
    \caption{(\textit{top}) Typical weekly pattern in the power demand data. Two exemplary states weakly corresponding to the weekly and daily pattern are displayed as well. (\textit{bottom})  Unusually low power demand during the summer results in a generally large model deviation compared to what is learned from the winter period. Additionally, the LSTM does not foresee the holidays which results in even larger prediction errors.}
    \label{fig:powerdemand_data}
\end{figure}

The Power demand dataset \cite{UCRArchive} contains a single signal which is the power consumption of a dutch research institute recorded every 15 minutes over a total period of one year. Several characteristics make this dataset interesting as a toy real world dataset. It contains both short term (daily) and long term (weekly) temporal patterns characterised by increased energy consumption during work days. It contains ``anomalies" in the form of holidays with significantly lower energy consumption (see Figure \ref{fig:powerdemand_data}). Additionally, there is a seasonal shift as the power demand decreases during the summer, only to increase again at the end of the year. 

We use the data from January 2.\ to March 21.\ as training data because the seasonal shift is slow enough and there are no holidays during this period. We will treat this problem analogous to a control system, where system input consists of two Boolean variables marking the beginning of each day and each weak. (Which is zero all the time except once a day/week, where it is one).
The task is to predict the current power consumption.

\begin{figure*}
    \centering
    \includegraphics[width=0.95\textwidth]{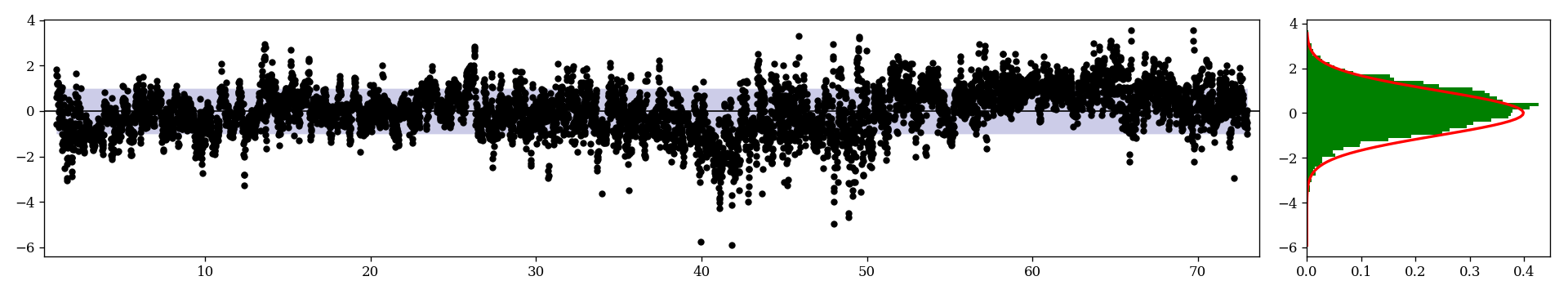}
    \includegraphics[width=0.95\textwidth]{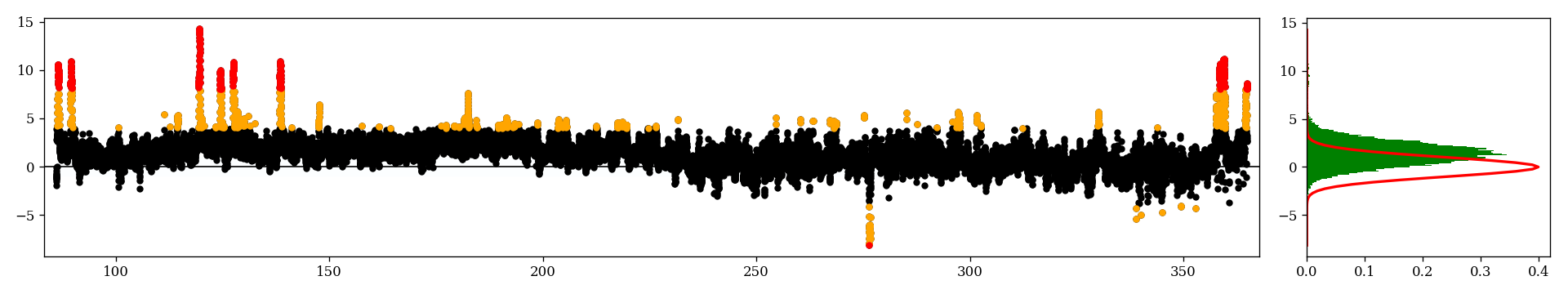}
    \caption{(\textit{Top}) Residuals of the power demand training set.\ (\textit{Bottom}) Residuals of the test set. The seasonal shift is noticeable in the first half of the data sample. Large positive model deviations indicate holidays. Points with absolute deviations greater than $4\sigma$ are coloured orange, deviations greater than $8\sigma$ are coloured red.}
    \label{fig:powerdemand_residuals}
\end{figure*}

Figure \ref{fig:powerdemand_data} shows the model predictions ($\pm1\sigma$-band) together with actual data from training and test set (during summer). Additionally, two of a total of 10 internal long-term memory states that are weakly associated with daily and weekly patterns are shown. About 3/10 states resembled a daily pattern, 4/10 weekly while the rest could not be associated with a weekly or daily pattern. The top of Figure \ref{fig:powerdemand_residuals} shows the residuals of the training set after convergence. The distribution can hardly be distinguished from a normal distribution which suggests a successful training stage.

\begin{figure}
    \centering
    \includegraphics[width=0.49\columnwidth]{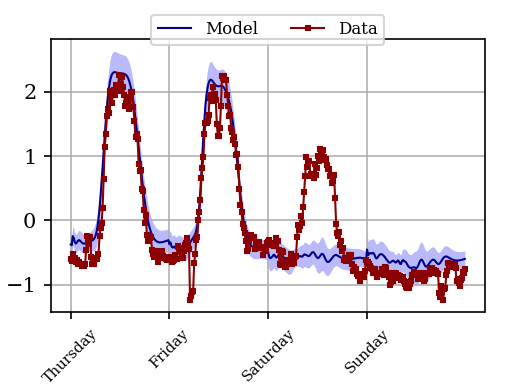}
    \includegraphics[width=0.49\columnwidth]{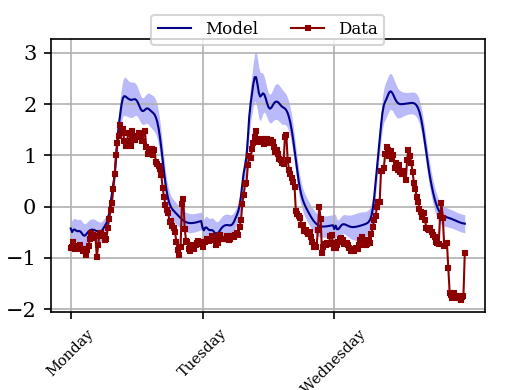}
    \caption{False positives of the open Saturday and End of year power drop.}
    \label{fig:falsePositives}
\end{figure}

The bottom of Figure \ref{fig:powerdemand_residuals} illustrates the anomaly detection approach.
To indicate the summer period we empirically choose a threshold of 4 standard deviations, while anomalies should exceed $8\sigma$ (without further optimisation). All holidays (days 87,90,120,125,128,139,359,360) have been detected with two additional false positives at day 365 (which exhibits a normal day pattern followed by a large power drop) and day 277 (which exhibits a pattern close to a working day, but on a Saturday), see Figure \ref{fig:falsePositives}.

%% file: conclusion.tex
\section{Discussion \& Remarks}\label{sec:discussion}

Another study of anomaly detection with the power demand data set can be found in \cite{malhotra2015long}. Here, the authors applied auto-regressive LSTM-sequence learning to the task, obtaining good results (0.94 precision, 0.17 recall and $F_{0.1}=0.90$). Precision is the fraction of anomalous data amongst all positively labelled instances, while recall is the fraction of detected anomalies. The $F_\beta$ score is the weighted harmonic mean of precision and recall, with $\beta$ weighing the importance of precision over recall. It should be noted that all observations on a holiday are to be labelled as anomalous, although most of the day (especially the night cycle) looks like any other day. This naturally leads to a high false positive rate for this problem and motivates to focus on a $F_\beta$ score with low $\beta$, putting a low weight on false positives.

There are a number of differences to our approach. First, we deliberately abdicate of the use of a validation set containing anomalies in order to fine-tune thresholds to make the model sensitive to the right kind of anomalies. Instead, as is motivated by real world scenarios where no anomalous behaviour has been observed, the threshold is chosen more or less arbitrary, by inspecting the distribution of normalised residuals of the training set (or alternatively of a small validation set).
Second, our approach is not auto-regressive but relies on input related to changes of system state. This makes the anomaly detection approach more robust to anomalous data (given also as input to the auto-regressive formalism before being compared to the model output), but less versatile to different kinds of data, where such input is not readily available or hard to construct given high-dimensional time-series data.
Using the aforementioned test-set as well as an $8\sigma$ anomaly threshold, final analysis results in a precision of 0.95, a 0.31 recall and $F_{0.1}=0.93$. Our method is sensitive to the missing power consumption on a holiday far earlier than the auto-regressive method simply by being independent of noisy anomalous model input, misleading the algorithm to assume a \textit{late rise} instead of a \textit{missing day}-pattern.
Because data is noisy, the positively sloped edge at the beginning of the shift occurs at slightly different times each day, which can be accounted for by proper uncertainty modelling.

The problem of model uncertainty, in particularly with regards to neural networks is still an open question of research, but definitely necessary to quantify in the domain of anomaly detection.
In this work, only the noise which is inherent in any data is modelled. Systematic uncertainty because of missing information or insufficient model complexity is not addressed (see \cite{gal2016uncertainty} for a general overview of uncertainty modelling in deep learning approaches).



Experiments showed that learning the variance along with the mean makes learning by gradient descent more difficult, even when using optimisers such as Adam \cite{kingma2014adam} or Momentum. It was beneficial to keep the $\tau$ variables frozen in a first learning stage.


The kind of data modelling described in this approach requires a  distinction of in- and output variables. Only the outputs are modelled and checked for anomalies. If anomalies are present in the inputs, the faulty system input should lead to a miss-prediction of output variables $\textbf{y}$ and therefore trigger the anomaly classification. However, this is not guaranteed and needs to be investigated in greater detail.
The Figures \ref{fig:watertank_anom_joined} and \ref{fig:watertank_anom_time_state} show, that the LSTM is able to extrapolate beyond the normally observed patterns in a meaning full way, indicating that even anomalous input could generate a ``correct'' (compared to the data) but nevertheless anomalous (given the context) output.

For many real world systems it is a problem to clearly distinguish between the $\mathbf{x}$ and $\mathbf{y}$ variables.
A simple partition into actor and sensor signals which is often possible in  control theory is not applicable to every system. 
One should keep in mind to always include all exogenous variables in $\mathbf{x}$, because they cannot be predicted by any other variable. In many CPS, some of these signals are binary in nature, such as on/off commands. Other possible candidates are environmental measurements that are not or only barely influenced by the system that is to be monitored.

A clear disadvantage of requiring a hard separation between in- and output variables is limited use in systems where such input can inherently not exist (e.g. heart rate monitoring or similar systems). Of course it is possible to adapt the model to include the output variables as input in something like an encoder-decoder recurrent neural network \cite{zhu2017deep}.
However, for the purpose of anomaly detection it is not guaranteed that every anomalous input is always going to generate an output significantly different to the observed value.


Although not discussed in this work, it is generally possible to use the approach described in this paper for time series forecast.
If the output prediction of observation vector $\mathbf{y}$ is to be estimated into the future one changes the training procedure for the LSTM to predict ahead of time: $f:(\mathbf{s}_{t-1},..,\mathbf{s}_{t+t'-1},\mathbf{x}_t)\mapsto (\mathbf{s}_{t+t'},\mathbf{y}_{t+t'})$ with $t'$ determining the shift into the future.

\section{Conclusions}\label{sec:conclusion}

We presented a model based anomaly detection for CPS data based on state dependent LSTM neural networks. 
In order to train the LSTM, a modified loss function intrinsically including the problem of noise prediction is derived. 
Equipped with this characteristic, LSTM are able to learn the temporal patterns inherent in CPS data and can output reliable predictions of observations made on the system. Predictions of both signal and noise can then be used to detect anomalies or system deterioration.